# An Enhanced YOLOv8 Model for Real-Time and Accurate Pothole Detection and Measurement

Mustafa YURDAKUL[1*], Şakir TAŞDEMİR[2]

[1]Kırıkkale University, Computer Engineering Department, Kırıkkale, Turkey, mustafayurdakul@kku.edu.tr , https://orcid.org/0000-0003-0562-4931

[2]Selçuk University, Computer Engineering Department, Konya, Turkey, stasdemir@selcuk.edu.tr , https://orcid.org/0000-0002-2433-246X

**Abstract**: Potholes cause vehicle damage and traffic accidents, creating serious safety and economic problems. Therefore, early and accurate detection of potholes is crucial. Existing detection methods are usually only based on 2D RGB images and cannot accurately analyze the physical characteristics of potholes. In this paper, a publicly available dataset of RGB-D images (PothRGBD) is created and an improved YOLOv8-based model is proposed for both pothole detection and pothole physical features analysis. The Intel RealSense D415 depth camera was used to collect RGB and depth data from the road surfaces, resulting in a PothRGBD dataset of 1000 images. The data was labeled in YOLO format suitable for segmentation. A novel YOLO model is proposed based on the YOLOv8n-seg architecture, which is structurally improved with Dynamic Snake Convolution (DSConv), Simple Attention Module (SimAM) and Gaussian Error Linear Unit (GELU). The proposed model segmented potholes with irregular edge structure more accurately, and performed perimeter and depth measurements on depth maps with high accuracy. The standard YOLOv8n-seg model achieved 91.9% precision, 85.2% recall and 91.9% mAP@50. With the proposed model, the values increased to 93.7%, 90.4% and 93.8% respectively. Thus, an improvement of 1.96% in precision, 6.13% in recall and 2.07% in mAP was achieved. The proposed model performs pothole detection as well as perimeter and depth measurement with high accuracy and is suitable for real-time applications due to its low model complexity. In this way, a lightweight and effective model that can be used in deep learning-based intelligent transportation solutions has been acquired.

## 1. Introduction

Potholes are one of the most common and dangerous types of road surface deterioration. It usually occurs when water seeps into the asphalt or concrete surface and weakens the sub-layers, then the traffic load erodes the weakened area[1, 2]. Over time, small cracks widen into deep potholes. Weather patterns, especially freeze-thaw cycles, accelerate this process, leading to rapid deterioration of the road surface[3]. In addition, heavy-tonnage vehicles further accelerate the formation of potholes by straining the bearing capacity of the asphalt.

Potholes not only deteriorate the road surface but also create serious risks to traffic safety. They can damage the vehicle suspension system, making it difficult to steer, and can cause tire punctures and rim warping[4]. Such deformations are life-threatening for motorcycle and bicycle riders and costly for cars and commercial vehicles. The presence of potholes slows down the flow of traffic, increasing fuel consumption and increasing travel times, resulting in economic losses[5, 6]. In busy city traffic, sudden braking and maneuvering caused by potholes can lead to accidents, endangering the safety of road users. For these reasons, early detection and repair of potholes is essential. Detecting potholes is not just about identifying their presence; an accurate measurement of their size, depth and volume ensures that the maintenance and repair process is managed efficiently. Without timely intervention, small deformations can grow rapidly, increasing repair costs and leaving road infrastructure vulnerable to damage on a larger scale. Deep potholes, in particular, can lead to water accumulation, rapidly causing greater damage to the road surface. Regular road inspections and continuous improvement of pothole detection systems are therefore a critical requirement for sustainable road maintenance management.

Traditional pothole detection methods rely on manual inspections by road maintenance crews. The process is usually based on visual observation, depth determination with mechanical measuring instruments and evaluation of driver complaints. However, these methods are time-consuming, labor-intensive, costly and prone to human error, as they involve a continuous process over a large road network. Periodic road inspections by road maintenance personnel can be inadequate, especially in large cities and on long stretches of highway. Furthermore, accurate measurement of critical parameters such as pothole depth, width and volume may not always be possible with manual methods.

These shortcomings highlight the need for computer-aided pothole detection systems. Artificial Intelligence (AI) and image processing technologies have made pothole detection faster, more accurate and cost-effective[7-10]. Laser Imaging Detection and Ranging (LIDAR) sensors, stereo cameras and radar systems can scan the road surface in three dimensions to accurately determine the size and depth of potholes[11, 12]. Deep Learning (DL) models can analyze large datasets to predict pothole occurrences and guide preventive maintenance efforts.

As a result, detecting and effectively managing potholes on road surfaces is critical not only to repair existing damage but also to ensure long-term infrastructure protection. Given the limitations of traditional methods, computer-aided solutions are more efficient, economical and sustainable road maintenance management.

The rest of the paper is organized as follows: Section 2 briefly reviews related work. Section 3 describes the proposed methodology, including data collection and model design. Section 4 reports the experimental results and comparisons. Section 5 discusses the findings and Section 6 concludes the paper.

## 2.Related Works

Studies on pothole detection and measurement on the road surface include a wide range of approaches, from traditional image processing methods to DL based approaches.

Vigneshwar and Kumar[13] preprocessed the image with differential gaussian filtering for pothole detection and then used various image segmentation methods. The edge detection method performed the best in pothole detection with 90.19% accuracy and 93.11% specificity. However, the K-Averages method stood out with the fastest processing time (0.27 s) and the highest sensitivity (87.18%). While edge detection was more successful in multiple pothole detection, overall the K-Averages method was a fast and effective alternative.

Lee et al.[14] compared background subtraction, convex hull, wavelet energy field, featured map, differential and otsu thresholding methods for pothole detection. The convex hull and wavelet energy field methods achieved the best results with 100% accuracy. The otsu thresholding method achieved 92.3% accuracy, while the background subtraction and differential methods underperformed with 76.9% accuracy.

Ouma and Hahn[15] used Fuzzy C-Means clustering (FCM) and morphological reconstruction. First, image filtering was performed with the wavelet transform, then pothole regions were grouped with FCM. Finally, morphological reconstruction was used to precisely define the contours of the potholes. The method achieved 87.5% Dice similarity coefficient, 77.7% Jaccard index and 97.6% sensitivity.

Akagić et al.[16] developed an unsupervised method based on image segmentation in RGB color space. First, the region was narrowed by detecting the asphalt surface, and then potholes were identified using Otsu thresholding. The algorithm was tested on 80 images from different camera angles and achieved 82% accuracy.

Garcillanosa et al.[17] used Canny edge detection and contour analysis-based image processing techniques with a Raspberry Pi microcontroller. The system performs pothole detection by acquiring images at a rate of eight frames/second with a camera mounted inside the vehicle. After the asphalt surface was identified by segmentation, pothole areas were detected by edge detection and contour analysis. Tested at speeds of 10-40 km/h, the system achieved 93.72% accuracy.

Ajay et al.[18] developed an image processing method based on morphological operations, Gaussian blur and Canny edge detection. First, video images were segmented into frames, then converted to grayscale to reduce noise. The potholes were highlighted with morphological operations and the pothole boundaries were identified with Canny edge detection. The proposed method achieved 77% accuracy.

Wang et al.[19] developed a method based on the wavelet energy field and Markov random field model. The wavelet energy field highlighted pothole regions by combining gray level and texture information. The potholes were then detected using morphological operations and geometric criteria. The Markov random field model was applied to precisely identify pothole edges, achieving 86.7% accuracy, 83.3% precision and 87.5% recall.

Hoang[20] developed an AI model using steerable filter, Gaussian filter and integral projection based feature extraction. In the study, pothole detection was performed using LS-SVM and Artificial Neural Network (ANN). The LS-SVM model provided the most successful results with 89% accuracy and 0.96 AUC score, while the ANN model achieved 85.25% accuracy and 0.92 AUC score.

While traditional image processing-based pothole detection methods are successful under certain conditions, they have serious limitations in real-world applications. These methods usually rely on techniques such as edge detection, thresholding, fuzzy clustering and color-based segmentation. However, they lose stability in the face of factors such as changing lighting conditions, shadows on the road surface and different shades of asphalt. Moreover, most approaches only work on 2D images, which makes them inadequate for pothole size estimation. On the other hand, with the increasing amount of data, traditional methods suffer from performance issues in large-scale data processing and cannot adapt to real-time applications.

At this point, DL-based approaches are emerging to overcome the shortcomings of traditional methods. Convolutional neural networks (CNN) and transformer-based segmentation models can automatically learn high-level features without the need for human intervention, and when trained on large data sets, they provide more consistent results under different environmental conditions.

Saisree et al.[21] used ResNet50, InceptionResNetV2 and VGG19 models to classify road surfaces as plain and pothole. According to the test results, the VGG19 model provided the highest accuracy rate of 97.91%.

Alzamzami et al.[22] used a YOLOv8-based detection model and unmanned aerial vehicle (UAV). The UAV scanned the road surfaces with an embedded camera system. To keep track of the detected potholes, a management interface for road maintenance workers and a web application showing pothole locations to drivers were developed. According to the test results, the YOLOv8 model achieved an f1-score of 95%, 98% accuracy and 92% recall rate.

Karukayil et al.[23] developed a system integrating a YOLO-based model and LIDAR sensors for pothole detection. The area, volume and depth of the pothole were calculated by combining RGB camera images with LIDAR data. YOLOv5, YOLOv6, YOLOv7 and YOLOv8 models were compared and YOLOv5 performed the best with 86% accuracy and 75.3% mAP@0.5.

Xing et al.[24] used Mask R-CNN and 3D reconstruction with binocular stereo vision. Pothole detection and segmentation was performed with Mask R-CNN on images acquired with a camera on the vehicle. Then, 3D pothole depth and surface area were calculated using binocular stereo vision. 98% detection accuracy and 94% segmentation success were achieved.

Bhavana et al.[25] proposed the POT-YOLO model for pothole detection. Frames were extracted from the video and preprocessed with Contrast Stretching Adaptive Gaussian Star Filter (CAGF), then pothole areas were identified with Sobel edge detection and classified with the proposed model. 99.1% accuracy, 97.6% precision, 93.52% recall and 90.2% f1-score were achieved. POT-YOLO provided higher accuracy compared to Faster R-CNN, SSD and Mask R-CNN models.

Talha et al.[26] developed a LIDAR and DL based system for pothole detection and size estimation. The proposed method analyzed the road surface by processing LIDAR data and performed pothole detection with the YOLOv5 model. The algorithm identifies pothole regions by giving 2D histograms obtained from LIDAR points as input to the YOLO model. The detected potholes were mapped by matching them with

Global Navigation Satellite Systems (GNSS) data. According to the experimental results, the YOLOv5s model achieved 96.5% accuracy, 94.8% precision and 95.3% recall.

In the existing literature, both traditional image processing techniques and DL based methods have been proposed for pothole detection. However, the vast majority of these studies only operate on 2D RGB images and therefore cannot accurately estimate the physical characteristics of the pothole. Moreover, the commonly used classification and object detection approaches only perform analysis at the bounding box level, which is insufficient for real measurement calculations.

Some studies that include depth information use sensors such as LIDAR or stereo cameras, but most of these systems rely on implicit source datasets. This imposes significant limitations in terms of both reproducibility and objective comparability of methods. Furthermore, these studies have both low detection performance and practical challenges in terms of applicability in real-world settings.

Although the methods proposed in the literature have made significant progress in pothole detection, they have several limitations in terms of physical measurement accuracy, data availability and suitability for real-time applications.

In this context, in light of these developments and existing shortcomings, the contributions of this study to the literature can be summarized as follows:

- A portable system capable of collecting real-time data from the road surface was developed.
- The first publicly available pothole dataset (PothRGBD) containing RGB and depth images and labeled for segmentation is introduced.
- An enhanced YOLOv8n-seg model integrating DSConv, SimAM, and GELU has been proposed to achieve accurate and morphologically robust segmentation.
- The proposed model is suitable for real-time and portable applications due to its low number of parameters and computational load.
- Accurate perimeter and depth estimations were obtained by combining the detected pothole regions with corresponding depth data.

These contributions collectively provide a comprehensive solution to the challenges observed in the existing literature and set the foundation for future developments in pothole detection systems.

## 3. Methodology

### 3.1. Dataset collection setup

A portable system was designed to collect depth and RGB image data of potholes from road surfaces. The system was built with the LattePanda 3 Delta computer with Windows operating system, Intel RealSense D415 depth camera, touch screen and powerbank. LattePanda 3 Delta provides a suitable computer platform for portable systems thanks to its compact structure, powerful processor and wide range of connectivity options.

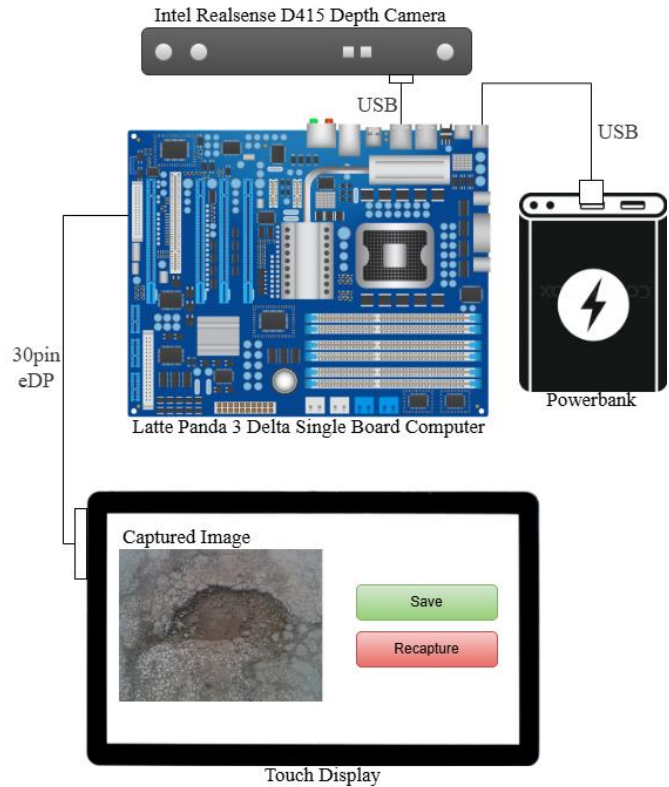

**Fig. 1** Schematic diagram of the pothole data collection system

The Intel RealSense D415 camera was chosen for pothole detection and dimension measurement, with the ability to obtain high-resolution RGB images and precise depth maps. The touchscreen provides instant feedback to the user, making it easy to view and manage data. The Powerbank enabled mobile use of the device, providing uninterrupted power for long-term data collection. Fig. 1 schematically shows the components and connectivity of the device developed for data collection. RGB and depth images were recorded simultaneously by walking on road surfaces.

### 3.2. Image dataset and annotation
Within the scope of the study, a total of 1000 data of 640x480 dimensions were collected and the resulting dataset was called Pothole RGB and Depth (PothRGBD) and released publicly. The labeling of the data was done in the Roboflow platform in line with the YOLO format and segmentation technique. In the experimental studies, 844 were used for training and 156 were used for testing. Fig. 2 shows randomly selected images from the PothRGBD dataset.

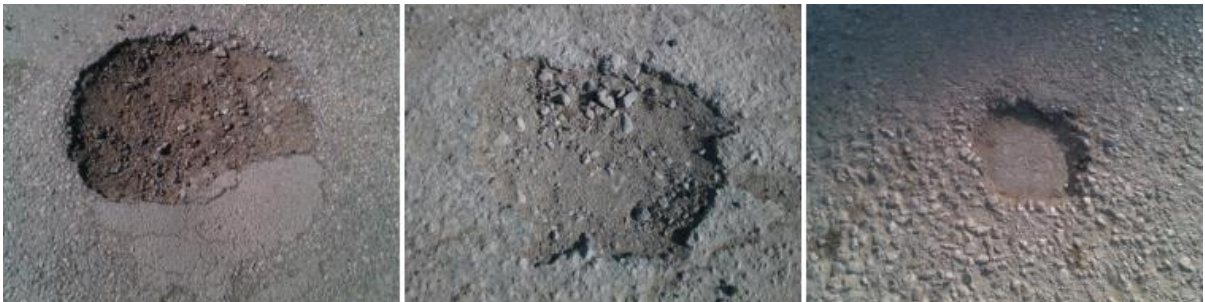

**Fig. 2.** Sample images from the PothRGBD dataset

### 3.3. Pothole measurement techniques

In real-world applications, if the system is mounted on a vehicle, the height of the camera will constantly vary depending on the suspension system and road conditions while the vehicle is in motion. This can lead to a loss of accuracy in pothole depth calculations. To determine the current height of the camera, a convenient approach is to calculate the median of the depth values of the pothole-free surface regions. In this way, the height variations caused by the dynamic movements of the vehicle will be eliminated and a more accurate depth measurement will be obtained by subtracting this reference height from the depth value in the region where the pothole is detected. In order to determine the camera height, we first divided the image depth into two clusters: pothole and non-pothole regions. Since the non-pothole regions represent the general level of the road surface, the median of the pixel depth values in this region is defined as $h_c$. The depth values in the regions containing potholes are determined by the segmentation model and expressed as $h_p$. Eq. 1 was implemented to obtain the actual pothole depth.

$$d = h_p - h_c \tag{1}$$

The segmentation output of the model is used to calculate the perimeter of each detected pothole. The model produces a mask marking only the region of the pothole. On this mask, the boundary pixels classified as "pothole" were followed along a string-like line. The coordinates of these pixels that form the boundary of the pothole in the segmentation mask are expressed as an ordered set as shown in Eq. 2.

$$P = \{(x_1, y_1), (x_2, y_2), \ldots, (x_n, y_n)\} \tag{2}$$

In Eq.2, $(x_i, y_i)$ represents the position of the ith pixel on the image at the boundary of the pothole region. These coordinates are in pixels and need to be converted into physical units. This transformation is done using the intrinsic parameters of the depth camera.

The overall perimeter length L in the real world is calculated by taking the sum of the physical distances between sequential boundary pixels, as shown in Eq. 3.

$$L = \sum_{i=1}^{n-1} \sqrt{[(x_{i+1} - x_i) \cdot s_x]^2 + [(y_{i+1} - y_i) \cdot s_y]^2} \tag{3}$$

In Eq. 3, $(x_i, y_i)$ represents the coordinates of the ith pixel at the pothole boundary obtained from the segmentation output. $s_x$ and $s_y$ denote the real physical length of a pixel on the horizontal and vertical axis, respectively, and are calculated using the intrinsic parameters of the depth camera. $n$ denotes the total number of pixels at the pothole boundary, and $L$ denotes the total real perimeter length. With this method, the perimeter length is accurately estimated by converting the pothole geometry determined by segmentation into real-world units of measurement.

### 3.4. Standart YOLOv8

The YOLO is known for its light weight, high performance and succesful outputs in many studies[27-33], In this study YOLOv8n [34, 35]was used to segment potholes in this study. The model consists of three basic structures: The main network (backbone) that extracts features, the intermediate network (neck) that combines features, and the detection module (head) that performs the prediction process. The main network extracts significant features at different scales from the images using convolutional methods. The intermediate network combines these features extracted by the main network to make them more meaningful. At this stage, feature pyramid networks (FPN) are usually preferred to effectively fuse lower-level information with higher-level information. In the detection module, images are classified and analyzed in detail thanks to detection receptors of different sizes. The CBS modules that form the basic structure of the model consist of convolution layer, stack normalization and SiLU activation function. C2f modules play a critical role in increasing the depth and quality of feature extraction by progressively analyzing data at different resolution levels[36]. Thus, the model can more accurately and reliably detect targets in images. Fig. 3 illustrates the standard YOLOv8-segmentation model architecture.

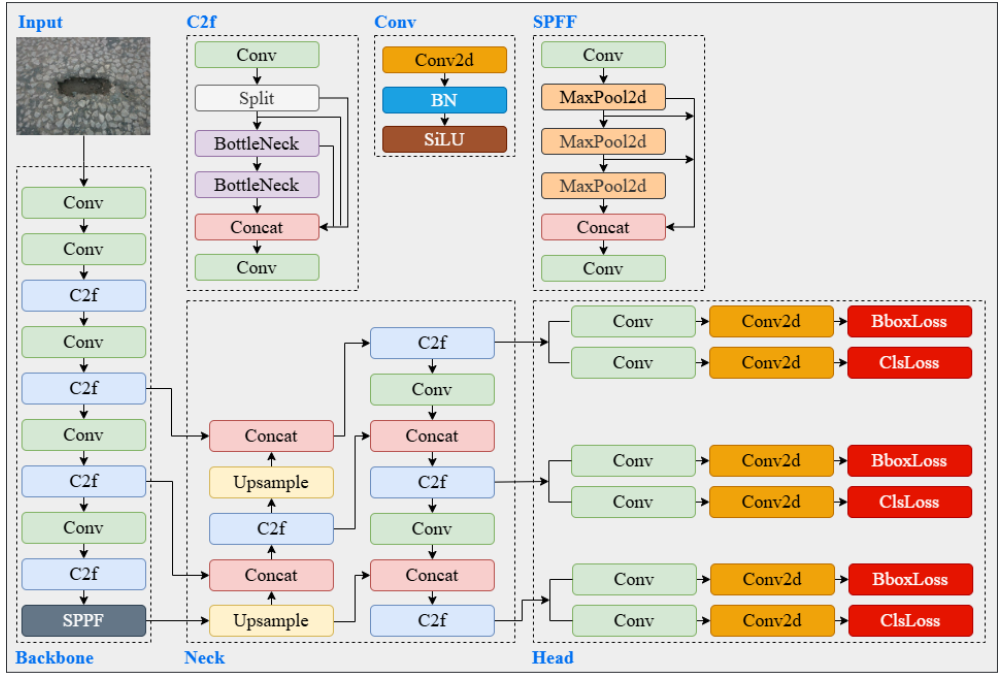

**Fig. 3** Schematic diagram of the standard YOLOv8 segmentation model architecture

### 3.5. Dynamic Snake Convolution

DSConv[37] is a geometry-aware convolution method developed for more accurate segmentation of thin and curved tubular structures. Instead of the traditional fixed grid structure, the method uses a sampling grid whose convolution kernel is dynamically aligned according to the local orientation of the target structure. The position of the grid points is defined by the cumulative offsets applied at each step starting from the kernel center. In both horizontal and vertical directions, this definition is expressed in Eq. 4.

$$K_{p\pm c} = \begin{cases} (x_p \pm c, y_p + \Sigma\Delta_y), \text{horizontal direction} \\ (x_p + \Sigma\Delta_x, y_p \pm c), \text{vertical direction} \end{cases} \quad (4)$$

$K_{p\pm c}$ is the grid position $c$ steps from the center of the kernel, $(x_p, y_p)$ is the center pixel coordinate of the kernel, $c$ is the distance from the center, $\Delta_x$ and $\Delta_y$ are the position offsets in $x$ and $y$ directions learned by the model. $\Sigma$ is the cumulative sum of offsets from previous steps. With this structure, the core samples parallel to the extension of the structure, allowing a better fit to the local morphology. However, since the computed positions usually correspond to fractional coordinates, the values at these points cannot be taken directly. This approach allows DSConv to focus the convolution kernel on the right regions, especially for curved and thin structures. In this way, more accurate feature extraction is achieved and the topological integrity of the structure is better preserved. Especially in regions where classical convolution methods are inadequate, DSConv produces stronger and more continuous segmentation results by aligning to the natural extension of the structure.

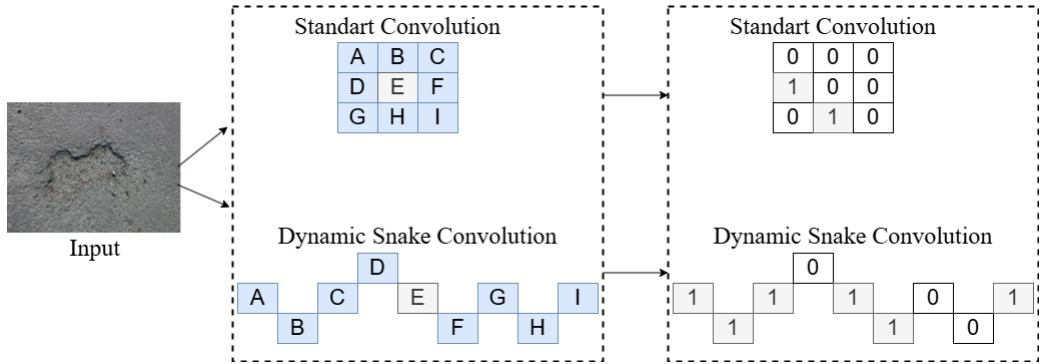

**Fig. 4** Comparison between standard convolution and DSConv showing convolution behavior over a pothole.

As can be seen in Fig. 4, the standard convolution kernel uses a fixed and limited sampling window to extract information only from a specific region of the structure, while the DSConv kernel focuses on more meaningful regions with a flexible grid structure shaped according to the orientation of the target structure. In this way, the resulting output feature maps provide both higher structural integrity and significantly improve segmentation accuracy by increasing morphological sensitivity.

### 3.6. GELU activation function

GELU[38] is an efficient activation function, particularly used in DL models such as transformers and modern CNNs. The input value x is scaled by multiplying its probability of occurrence under a standard normal distribution by $\phi(x)$. GELU combines the linear bias of ReLU and the smooth, probabilistic character of the sigmoid. It passes the signal at high values and softly suppresses the transition at low and negative values.

$$GELU(x) = x \cdot \Phi(x) \tag{5}$$

Studies in the literature have shown that GELU provides better gradient flow, reduces convergence time and improves feature extraction. In object detection models such as YOLOv8s, using GELU can improve overall performance by helping to recognize finer details. SiLU also exhibits similar performance; however, GELU's distribution-based structure has the potential to yield better results, particularly for data containing complex patterns.

### 3.7. SimAM

SimAM[39] is a parameter-free and lightweight attention mechanism that determines the importance of each neuron by measuring how outlier it is relative to its surroundings. The module generates attention weights directly in 3D and does so through an energy function. The energy defined for each neuron expresses how different that neuron is from the channel average and is calculated by the formula in Eq. 6.

$$e_t = \frac{4(\sigma^2 + \lambda)}{(x_t - \mu)^2 + 2\sigma^2 + 2\lambda} \tag{6}$$

Where $x_t$ is the value of the target neuron, $\mu$ and $\sigma^2$ are the within-channel mean and variance, respectively, and $\lambda$ is a fixed compensation parameter. As the energy value decreases, the importance of the neuron increases. This energy value is converted into an attention coefficient to calculate the output of the neuron.

$$\hat{x}_t = sigmoid\left(\frac{1}{e_t}\right) \cdot x_t \tag{7}$$

In this way, more outlier and information-carrying neurons are highlighted. Since SimAM does not require extra parameters, it is both efficient and easily integrated into existing networks.

### 3.8. Proposed YOLOv8 Model

The classical YOLOv8-seg architecture was redesigned to more accurately detect potholes, which are irregular in shape, have degraded edge structure, and are difficult to distinguish from context. The most fundamental change is the replacement of certain Conv blocks in the backbone and head layers with DSConv. In contrast to the fixed grid structure of conventional convolutions, DSConv is able to adaptively update kernel positions with learnable deformations, allowing it to track curved, broken or irregular edges much more accurately. This feature allows the model to better learn the degraded structures such as cracks, potholes and similar deformations, especially on asphalt. The main reason for not intervening in all layers is to keep both the number of parameters and the computational load of the model to a minimum, while maintaining real-time functionality. Since the first Conv layer in the input only extracts low-level features such as basic edge and color transition, DSConv is not used in this part. In addition, in order to strengthen the attention mechanism of the model, a parameter-free and lightweight SimAM was added to the output of all C2f blocks in the backbone and neck region. SimAM evaluates the importance of each pixel using a neural energy approach, allowing the model to focus on more meaningful areas. This module was particularly effective in the neck section, where multi-scale information fusion was performed, and contributed to a more careful blending of different levels of detail. However, the attention module was not added to early stage layers, such as the first C2f block, because these layers have not yet started to learn discriminative features and using attention at this level could potentially impair the learning process. Finally, the activation function in all Conv blocks was replaced by GELU instead of the default SiLU. The smooth transition and low gradient loss structure of GELU makes the learning more stable and makes it work harmoniously with the attention and deformation modules. Thus, the proposed architecture has been improved in terms of both accuracy and geometric precision, while maintaining the lightweight structure of the original model. A schematic diagram of the enhanced YOLOv8 segmentation model is shown in Fig. 5.

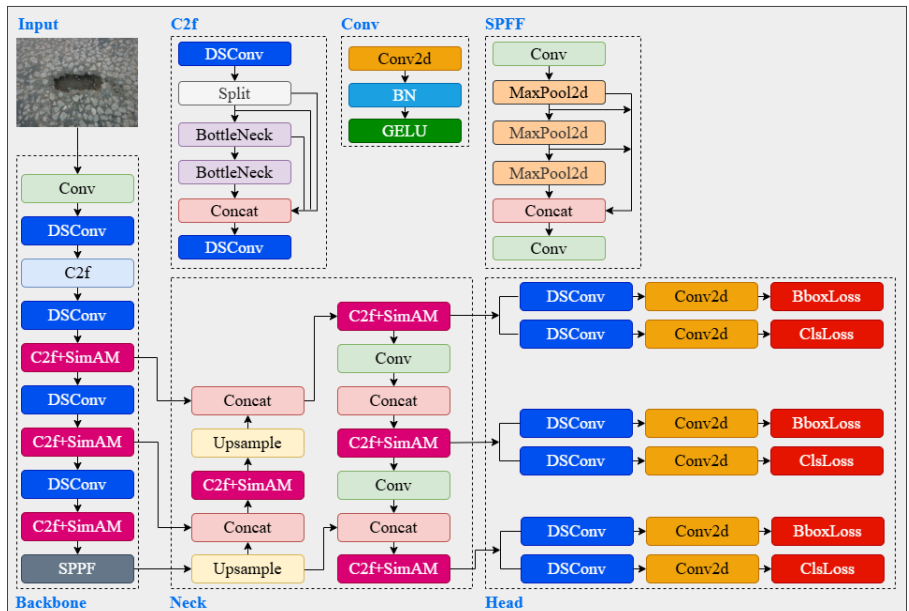

**Fig. 5** Schematic representation of the enhanced YOLOv8 architecture with DSConv, SimAM, and GELU integrations.

### 3.9. Evaluation Metrics

The model's performance is evaluated using the metrics precision, recall, Mean Average Precison (mAP) and Floating Point Operations per Second (FLOP). Precision indicates how many instances the model correctly detects as positive, while recall indicates how many true positives are detected. mAP reflects the average detection success at different thresholds. FLOPs indicate the computational load of the model. The mathematical formulas of the metrics are given in Eq.8-11. TP (True Positive) refers to correctly detected positives; FP (False Positive) refers to false positives; FN (False Negative) refers to missed positives; TN (True Negative) refers to correctly excluded negative samples. $C_{in}$ and $C_{out}$ are the number of input and output channels of the layers; $K_H$ and $K_W$ are the dimensions of the kernel; $H_{out}$ and $W_{out}$ are the height and width of the output map.

$$Precision = TP/(TP + FP) \tag{8}$$

$$Recall = TP/(TP + FN) \tag{9}$$

$$mAP = \frac{1}{N}\sum_{i=1}^{N} AP_i \tag{10}$$

$$FLOPs = 2 x C_{in} x H_{out} x W_{out} x K_H x K_w x C_{out} \tag{11}$$

## 4.Results

The experiments in this study were performed on a high-performance computer system. The operating system used was Windows 11 and Python 3.11.4 was chosen as the programming language. TensorFlow 2.14.0, Ultralytics 8.3.121, Pytorch 2.5.1 and Matplotlib 3.7.1 libraries were used to develop and visualize the segmentation models. The computer is equipped with two Nvidia RT3090 GPUs (24GB each), Intel Core i9-10920X CPU (3.50GHz) and 128GB RAM. In addition, CUDA version 12.7 was used for GPU accelerated operations.

### 4.1. Ablation Experiment
Ablation experiments were conducted in this section to evaluate the contributions of structural improvements applied to the YOLOv8n-seg baseline model to performance. Each modification was applied as an independent experiment, repeated five times, and the results were reported as the mean and standard deviation to allow for a statistically reliable comparison with the baseline model. Table 1 summarizes the average performance metrics along with their standard deviations.
The baseline model, YOLOv8n-seg, achieved 91.9% accuracy, 85.2% recall, and 91.9% mAP@50. The model's number of parameters was measured as 3.2M, its computational load as 12.1G FLOPs, and its Frames Per Second (FPS) value as 121.
In the first experiment, specific convolution blocks in the baseline model were replaced with DSConv. As a result of this modification, accuracy was 92.7%, recall was 86.1%, and mAP@50 was 92.6%. Compared to the baseline model, accuracy increased by 0.8%, recall by 0.9%, and mAP@50 by 0.7%. The number of parameters increased from 3.2M to 4.1M, the FLOPs value increased from 12.1G to 13.0G, and the FPS value decreased from 121 to 115.
In the second experiment, the SimAM attention mechanism was added to highlight regions that carry information to specific parts of the baseline model. With this modification, accuracy reached 92.4%, recall reached 86.3%, and mAP@50 reached 92.4%. Compared to the baseline model, there was an increase of 0.5% in accuracy, 1.1% in recall, and 0.5% in mAP@50. The number of parameters remained unchanged (3.2M), and the FPS value was 119.
In the third experiment, the activation function was adjusted in the baseline model; GELU, which enables more stable learning, was used instead of the default SiLU in all convolution layers. As a result of this modification, accuracy was achieved 92.2%, recall 85.9%, and mAP@50 92.2%. Compared to the baseline model, there was an increase of 0.3% in accuracy, 0.7% in recall, and 0.3% in mAP@50. There was no change in the number of parameters, and the FPS value was measured as 120.

Then, DSConv and SimAM configurations were applied together on the baseline model, achieving accuracy of 93.1%, recall of 87.0%, and mAP@50 of 93.3%. The number of parameters was 4.1M, the FLOPs value was 13.2G FLOPs, and the FPS value was 112. In the final stage, DSConv, SimAM, and GELU modifications were applied together to the baseline model, and the proposed final model was created. In this model, accuracy was obtained 93.7%, recall 90.4%, and mAP@50 93.8%. The number of parameters is 4.1M, FLOPs is 13.2G FLOPs, and the FPS value is 110.

When compared to the baseline YOLOv8n-seg model, the final model achieved an increase of 1.8% in accuracy, 5.2% in recall, and 1.9% in mAP@50. Additionally, there was an increase of +0.9M in the number of parameters, +1.1G in FLOPs, and a decrease of 11 units in FPS. However, these increases in computational load are easily tolerable by today's GPU architectures or efficient edge devices. Therefore, the resulting load increase is a device-dependent factor and does not pose a practical obstacle in real-time applications. Additionally, low standard deviations across different configurations show that the improvements with DSConv, SimAM, and GELU not only provide higher accuracy but also stable and consistent performance. It indicates that the model generalizes well and is less sensitive to training variations.

**Table 1.** Ablation study results showing the contribution of each component (DSConv, SimAM, GELU) to the YOLOv8n-seg baseline model

| Model | Performance evaluation metrics | | | | | |
| --- | --- | --- | --- | --- | --- | --- |
|  | P/% | R/% | mAP%50 | Params/M | FLOPs/G | FPS |
| YOLOv8n-seg (Baseline) | 91.9±0.4 | 85.2±0.5 | 91.9±0.5 | **3.2** | **12.1** | **121** |
| +DSConv | 92.7±0.3 | 86.1±0.5 | 92.6±0.4 | 4.1 | 13.0 | 115 |
| +SimAM | 92.4±0.3 | 86.3±0.3 | 92.4±0.5 | 3.2 | 12.3 | 119 |
| +GELU | 92.2±0.2 | 85.9±0.5 | 92.2±0.3 | 3.2 | 12.1 | 120 |
| +DSConv + SimAM | 93.1±0.4 | 87.0±0.5 | 93.3±0.4 | 4.1 | 13.2 | 112 |
| **+DSConv + SimAM + GELU (Proposed)** | **93.7±0.3** | **90.4±0.4** | **93.8±0.4** | 4.1 | 13.2 | 110 |

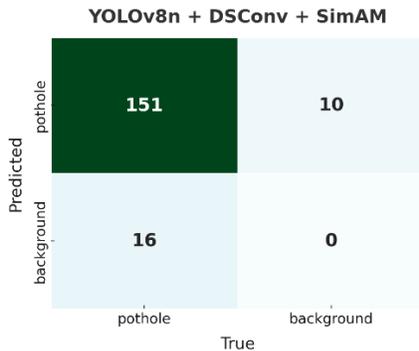

**Fig. 6** Average confusion matrix of the proposed model on the PothRGBD test data

Additionally, the classification performance of the proposed model was evaluated using the average confusion matrix obtained from five independent runs, as shown in Fig. 6. From a total of 161 pothole samples, 151 were correctly classified, and only 10 samples were labeled as false positives. In contrast, 16 potholes were incorrectly predicted as background, and no true positive examples were detected in the background class. The results indicate that the model accurately identifies the pothole class; however, due to the limited number of examples in the background class, it can only generalize to a limited extent. The high precision and recall values confirm that the model performs well, particularly in the target object class, which is the pothole class.

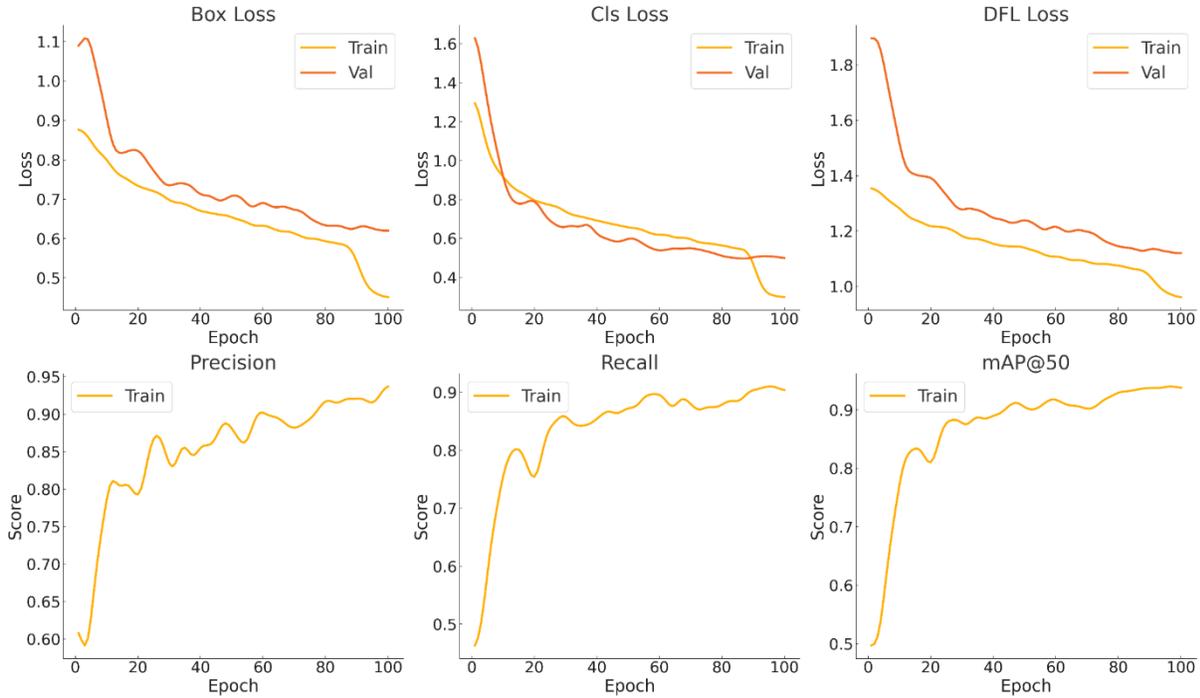
**Fig. 7** Training and validation performance of the proposed model across loss and evaluation metrics.

Fig. 7 shows the average trends in box loss, class loss, DFL (Distribution Focal Loss) values, and precision, recall, and mAP@50 metrics across five independent runs during the training and validation processes of the proposed model. A rapid decrease in all loss values was observed within the first 20 epochs, with box loss and class loss values also showing a steady decline in the validation set. It indicates that the model is able to generalize without showing overfitting.

Looking at the success metrics, a significant increase in precision and recall values was achieved, especially within the first 30 epochs. As the training process progressed, the curves of these metrics stabilized and followed a similar trend in the validation set. The mAP@50 value reached a high level at approximately the 40th epoch and stabilized after the 90th epoch. It shows that the model achieved a balanced performance in terms of both bounding box quality and classification success.

### 4.2. Comparative analysis of segmentation performance utilizing various models

In this section, the segmentation performance of the proposed model is compared with the YOLOv8 (n, s, m), YOLOv9 (c, e), and YOLOv11 (n, s, m) models. For a comprehensive evaluation, different model generations and scales were evaluated. The experimental results are presented in Table 2. All performance values were computed as the mean and standard deviation over five independent training runs, ensuring the reliability of the comparisons. The YOLOv8n model achieved 91.9% accuracy and 121 FPS with only 3.2 million parameters, demonstrating high speed and low model complexity; however, it was limited to 85.2% recall. The YOLOv8s model achieved higher recall (87.6) and mAP (92.3) values, but showed a decrease in accuracy (88.6) and speed (60 FPS). YOLOv8m achieved the highest accuracy rate (92.1%) in its model family, but is notable for its low recall rate (83.0%) and high computational load (27.2 million parameters, 110.0G FLOP). YOLOv9c and YOLOv9e models have high parameters (27.8 million and 59.6 million, respectively) and extremely high FLOPs (159.1G and 244.4G). Despite its high mAP@50 value (92.4), YOLOv9e is not suitable for real-time applications due to its low FPS values (27 and 18). YOLOv11n achieves 90.4% recall and 111 FPS with only 2.8 million parameters and 10.2 G FLOP. However, its 88.4% accuracy value lags behind other models in terms of accuracy. While the YOLOv11s model demonstrates balanced performance (accuracy 88.9%, recall 87.0%), the YOLOv11m model exhibits the lowest accuracy value (83.5%). In contrast, the proposed model achieved the highest overall performance with 93.7% accuracy, 90.4% recall, and 93.8% mAP@50. It has 4.1 million parameters, 13.2 G FLOP, and 110 FPS. Compared to the YOLOv8n (3.2 million parameters, 12.1 G FLOP, 121 FPS) and YOLOv11n (2.8 million parameters, 10.2 G FLOP, 111 FPS) models, these values indicate slightly higher computational load and lower FPS.

**Table 2.** Comparison of segmentation performance metrics across different YOLO models (YOLOv8, YOLOv9, YOLOv11, and proposed model).

| Model | Performance evaluation metrics | | | | | |
|---|---|---|---|---|---|---|
| | P/% | R/% | mAP%50 | Params/M | FLOPs/G | FPS |
| YOLOv8n | 91.9±0.4 | 85.2±0.4 | 91.9±0.5 | **3.2** | **12.1** | **121** |
| YOLOv8s | 88.6±0.4 | 87.6±0.5 | 92.3±0.4 | 11.7 | 42.4 | 60 |
| YOLOv8m | 92.1±0.3 | 83±0.7 | 91.8±0.6 | 27.2 | 110 | 39 |
| YOLOv9c | 90.5±0.5 | 84.2±0.6 | 90.8±0.5 | 27.8 | 159.1 | 27 |
| YOLOv9e | 88.9±0.4 | 85.2±0.5 | 92.4±0.5 | 59.6 | 244.4 | 18 |
| YOLO11n | 88.4±0.5 | 90.4±0.5 | 91.9±0.6 | 2.8 | 10.2 | 111 |
| YOLO11s | 88.9±0.7 | 87±0.6 | 90.7±0.7 | 10 | 35.3 | 67 |
| YOLO11m | 83.5±0.7 | 86.7±0.5 | 88.7±0.5 | 22.3 | 123 | 36 |
| **Proposed** | **93.7±0.3** | **90.4±0.4** | **93.8±0.4** | 4.1 | 13.2 | 110 |

However, the difference is an acceptable trade-off for the accuracy gains provided by structural improvements such as DSConv, SimAM, and GELU. Additionally, today's modern GPUs and efficient embedded systems can easily tolerate this additional load. Additionally, optimization techniques (quantization, TensorRT, model pruning, etc.) are available to increase the FPS value and make the model usable on devices with lower resources. In conclusion, the proposed model is a robust alternative for real-time and resource-constrained applications, balancing high accuracy with low computational cost. On the other hand, the consistently low standard deviations across all YOLO models demonstrate that the PothRGBD dataset is not only clean and consistently labeled, but also well suited for the segmentation task. The consistency across models highlights the stability and clarity of segmentation boundaries in the dataset and ensures that even basic architectures perform reliably across multiple experiments.

### 4.3. Measurement results

The physical measurement performance of the proposed model was evaluated by analyzing sample images. For each image, the model's predictions were compared with the actual environment and depth values of the pothole, and the difference values were calculated and presented in Table 3.

In the comparison conducted using five images, the differences in perimeter predictions ranged from -3.2 cm to +1.7 cm, with an average difference of approximately ±2.3 cm. In depth predictions, the differences ranged from -0.2 cm to +0.3 cm, with an average error of approximately ±0.24 cm.

When evaluated in absolute terms, these differences are considered small. Particularly in field applications, a deviation at this level is not considered a critical error in terms of road maintenance prioritization. A difference of ±0.3 cm in depth or ±2–3 cm in perimeter does not pose a significant risk in real-world applications and does not compromise the reliability of the system.

**Table 3.** Quantitative comparison of real versus predicted pothole measurements using the proposed approach.

| Image | Real(cm) | | Proposed Approach(cm) | | Difference(cm) | |
|---|---|---|---|---|---|---|
| | Perimeter | Depth | Perimeter | Depth | Perimeter | Depth |
| 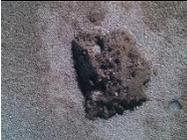 | 127.6 | 6.2 | 125.1 | 6.0 | -2.5 | -0.2 |
| 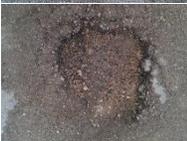 | 96.3 | 4.8 | 97.9 | 5.0 | +1.6 | +0.2 |
| 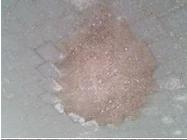 | 104.2 | 5.5 | 101.7 | 5.3 | -2.5 | -0.2 |
| 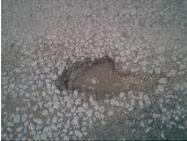 | 88.5 | 3.9 | 90.2 | 4.2 | +1.7 | +0.3 |
| 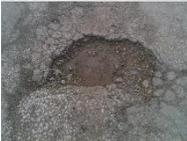 | 144.8 | 5.4 | 141.6 | 5.7 | -3.2 | +0.3 |

In particular, in perimeter measurement, the conversion of contours extracted from the segmentation mask into physical units reflects the morphological accuracy of the model. In depth measurement, the proposed reference ground plane method successfully balances changes in camera height.

## 5.Discussion

In this study, an enhanced YOLO segmentation model was developed to detect potholes on road surfaces and calculate their physical characteristics (perimeter and depth) for real-time applications. The YOLOv8n-seg architecture was used as a basis for model development, and a systematic ablation analysis was conducted to evaluate the contribution of structural improvements. In this context, DSConv, SimAM, and GELU components were integrated into the model, and the effect of each improvement on segmentation performance was analysed quantitatively. The results show that each module contributes to the model and that the combined use of the components achieves high performance metrics of 93.7% precision, 90.4% recall, and 93.8% mAP@50. These performance metrics represent the average values of five independent studies with low standard deviations, demonstrating consistent and reliable results. Additionally, the proposed model was compared with YOLOv8, YOLOv9, and YOLOv11 versions; despite its low parameter count and computational load, it demonstrated the best performance in terms of accuracy-efficiency balance.

The segmentation outputs of the model on the test data are presented in Fig. 8. In examples 1, 3, 6, 10, 11, and 15 in the figure, the model accurately segments the potholes; it can successfully distinguish edge details, especially thanks to the DSConv module's adaptation to curved structures. In examples 2, 4, 5, 7, 8, and 16, the segmentation is generally correct, but some edge regions show overflows or missing coverage. In these cases, the model is uncertain at the boundaries because the potholes and the road surface have similar textures and colour tones. In images 9, 12, 13, and 14, the model is observed to struggle more. In such scenes, the presence of multiple potholes or the complexity of surface patterns makes it difficult for the model to focus and clearly define pothole boundaries.

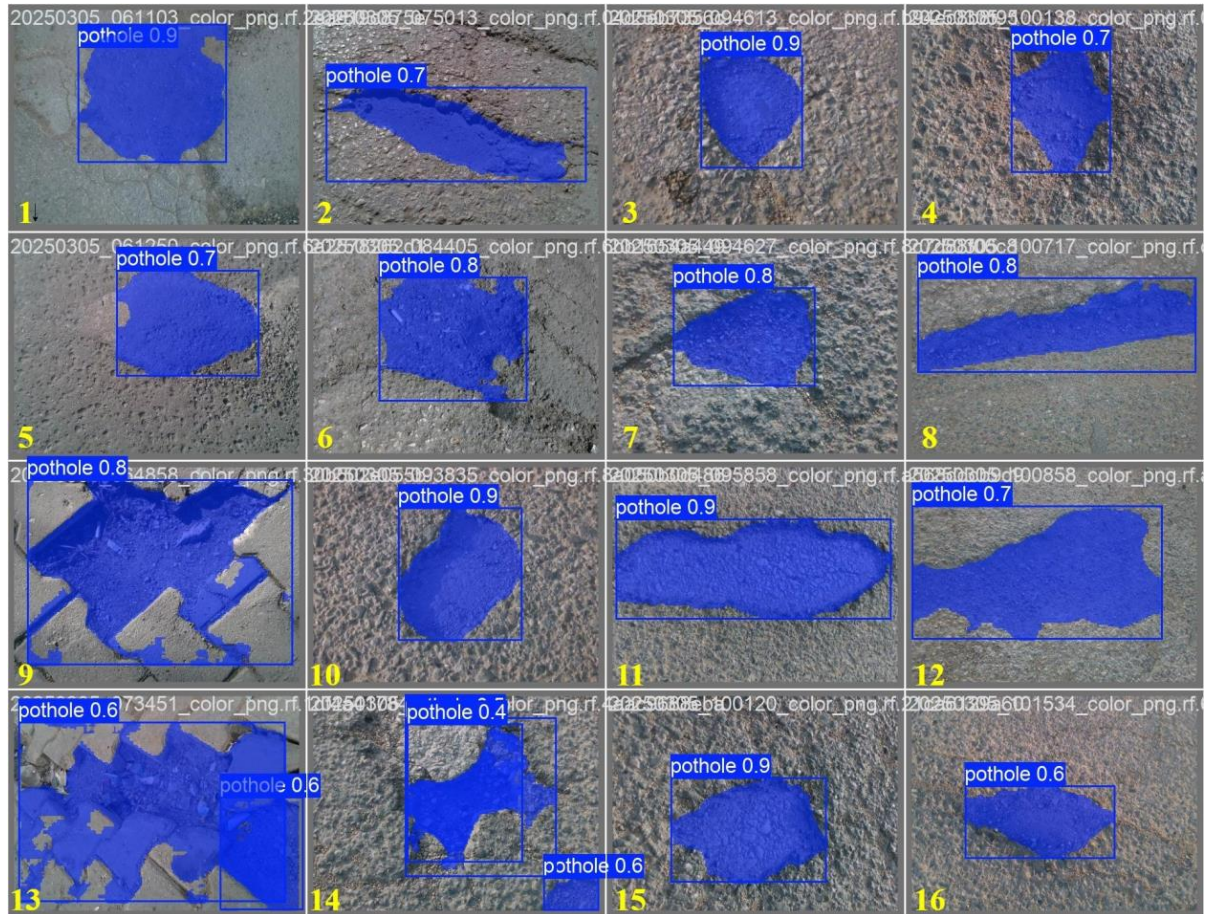

**Fig. 8** Sample segmentation results of the proposed model on the PothRGBD test set.

In addition to segmentation performance, the effect of the model's attention mechanism was also evaluated through Grad-CAM-like visual attention maps. Fig. 9 shows the contributions of the proposed model (YOLOv8+DSConv+SimAM) in terms of attention mechanism, compared with the original YOLOv8 and the version with only DSConv integration. When looking at the visual attention maps, although the YOLOv8 model is able to cover the pothole area in general, it is observed that the attention distribution is quite wide and spread out. The DSConv-based structure centralises the attention area, increasing focus on the target object and thus showing more stable attention performance. The most successful results were obtained with the YOLOv8-DSConv+SimAM model. This model offers a more selective and consistent attention distribution with both high focus clarity and low levels of environmental noise in the attention maps. The model is thus able to concentrate solely on the target object, enabling more accurate and reliable detections.

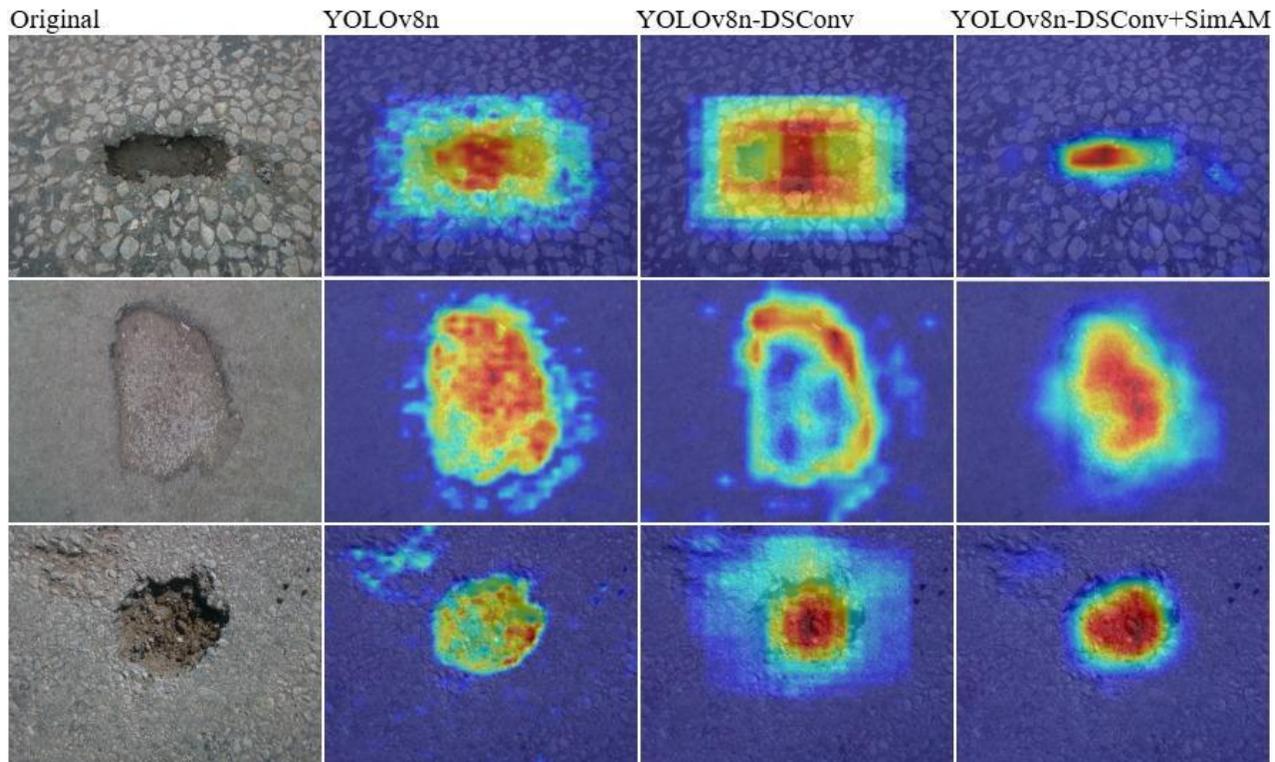

**Fig. 9.** Attention heatmap comparison of segmentation models: standard YOLOv8n, YOLOv8n + DSConv, and

These findings demonstrate that the proposed model has been structurally optimized and is suitable for practical applications. The developed system is considered to be effective for tasks such as pothole detection and reporting in embedded systems, unmanned aerial vehicles, or vehicle-mounted control systems. However, it has been observed that the model experiences uncertainty in boundary determination on certain ground types. In such cases, the model's classification performance is limited due to low contrast between classes.

Future studies plan to expand training with datasets that include more diverse ground types, investigate multi-attention modules, and statistically validate physical measurement results with a larger field dataset. Additionally, studies aimed at improving measurement accuracy through post-processing steps after segmentation can be conducted. Systems developed in this direction have the potential to evolve into a digital decision support infrastructure that can be directly integrated into urban road maintenance planning.

## 6. Conclusion

In this study, a new dataset (PothRGBD) consisting of RGB-D images has been created to accurately detect potholes and perform physical measurements. The dataset was collected from road surfaces using an Intel RealSense D415 depth camera, and 1,044 images were labeled in YOLO segmentation format. Thus, the first pothole dataset reflecting real-world conditions, containing depth information, and available in a publicly accessible repository has been proposed. The model architecture was developed based on YOLOv8n-seg, with the DSConv, SimAM, and GELU components, enabling more accurate segmentation of irregular and curved pothole structures. The experimental results, averaged from five independent runs, demonstrate the proposed model's high performance with high accuracy of 93.7%, recall of 90.4%, and mAP@50 of 93.8% with low standard deviations, indicating the consistency and statistical reliability of the results. The model demonstrated its suitability for real-time applications by running at 110 FPS with only 4.1 million parameters and 13.2 G FLOP. In comparative analyses, the proposed model showed higher accuracy than different versions of the YOLOv8, YOLOv9, and YOLOv11 families despite having lower model complexity. Only the YOLOv8n and YOLOv11n models have lower values in terms of parameter count and FLOPs; however, these models lag behind in terms of segmentation accuracy. The increase in computational load in the proposed architecture is easily tolerable by modern hardware and can be further improved with optimization techniques if necessary. Additionally, the proposed work not only detects potholes but also accurately estimates their surroundings and depth using segmentation masks. The proposed ground plane approach successfully compensates for varying camera heights in field conditions and improves measurement accuracy. In conclusion, this study has made a significant contribution to the field of pothole detection and measurement by providing both a unique dataset and a high-accuracy, low-cost segmentation model. Future work aims to expand the dataset with more diverse surface types, improve accuracy through post-processing, and integrate the model into real-time embedded systems.

**Data Availability:**
The PothRGBD dataset introduced in this study can be accessed following links:
https://github.com/ymyurdakul/PData
https://www.kaggle.com/datasets/mahyeks/pothrgbd-rgb-and-depth-images-of-potholes

**Funding:**
This research received no external funding.

**Competing Interests:**
The authors declare that they have no competing interests.

**Author Contributions:**
Conceptualization, M.Y. and Ş.T.; methodology, M.Y.; software, M.Y.; validation, M.Y. and Ş.T.; formal analysis, M.Y.; investigation, M.Y.; resources, Ş.T.; data curation, M.Y.; writing—original draft preparation, M.Y.; writing—review and editing, Ş.T.; visualization, M.Y.; supervision, Ş.T.; project administration, Ş.T. All authors have read and agreed to the published version of the manuscript.